%% file: main.tex
\documentclass[sigconf]{acmart}
\usepackage{algorithm}
\usepackage{algpseudocode}
\usepackage{amsmath}
\usepackage{graphics}
\usepackage{epsfig}
\usepackage{microtype}
\usepackage{graphicx}
\usepackage{subfigure}
\usepackage{booktabs}
\usepackage{hyperref}
\usepackage{float}
\renewcommand{\algorithmicrequire}{\textbf{Input:}} 
\renewcommand{\algorithmicensure}{\textbf{Output:}} 

\input{math_commands.tex}

\usepackage{multirow}
\newcommand{\real}{\mathbb{R}}

\newcommand{\ourSimpleModel}{SCE}
\newcommand{\ourMultiModel}{MoSCE}
\usepackage{stfloats}

\AtBeginDocument{%
  \providecommand\BibTeX{{%
    \normalfont B\kern-0.5em{\scshape i\kern-0.25em b}\kern-0.8em\TeX}}}

\copyrightyear{2020}
\acmYear{2020}
\setcopyright{acmcopyright}\acmConference[KDD '20]{Proceedings of the 26th ACM SIGKDD Conference on Knowledge Discovery and Data Mining}{August 23--27, 2020}{Virtual Event, CA, USA}
\acmBooktitle{Proceedings of the 26th ACM SIGKDD Conference on Knowledge Discovery and Data Mining (KDD '20), August 23--27, 2020, Virtual Event, CA, USA}
\acmPrice{15.00}
\acmDOI{10.1145/3394486.3403068}
\acmISBN{978-1-4503-7998-4/20/08}

\begin{document}
\fancyhead{}
\title{SCE: Scalable Network Embedding from Sparsest Cut}


\author{Shengzhong Zhang}
\affiliation{
\institution{Fudan University}}
\email{szzhang17@fudan.edu.cn}

\author{Zengfeng Huang}
\affiliation{
\institution{Fudan University}}
\email{huangzf@fudan.edu.cn}
\authornote{Corresponding author}

\author{Haicang Zhou}
\affiliation{
\institution{Fudan University}}
\email{haicang.zhou@outlook.com}

\author{Ziang Zhou}
\affiliation{
\institution{Fudan University}}
\email{zhouza15@fudan.edu.cn}

\renewcommand{\shortauthors}{anonymous}

\begin{abstract}
Large-scale network embedding is to learn a latent representation for each node in an unsupervised manner, which captures inherent properties and structural information of the underlying graph. In this field, many popular approaches are influenced by the skip-gram model from natural language processing. Most of them use a contrastive objective to train an encoder which forces the embeddings of similar pairs to be close and embeddings of negative samples to be far. A key of success to such contrastive learning methods is how to draw positive and negative samples. While negative samples that are generated by straightforward random sampling are often satisfying, methods for drawing positive examples remains a hot topic.

In this paper, we propose SCE for unsupervised network embedding only using negative samples for training. Our method is based on a new contrastive objective inspired by the well-known sparsest cut problem. To solve the underlying optimization problem, we introduce a Laplacian smoothing trick, which uses graph convolutional operators as low-pass filters for smoothing node representations. The resulting model consists of a GCN-type structure as the encoder and a simple loss function. Notably, our model does not use positive samples but only negative samples for training, which not only makes the implementation and tuning much easier, but also reduces the training time significantly. 

Finally, extensive experimental studies on real world data sets are conducted. The results clearly demonstrate the advantages of our new model in both accuracy and scalability compared to strong baselines such as GraphSAGE, G2G and DGI. 
\end{abstract}





\maketitle

\input{intro}

\input{Analysis}

\input{model}
\input{related}
\input{exp}
\input{conclusion}
\bibliographystyle{ACM-Reference-Format}
\bibliography{main}


\end{document}

%% file: math_commands.tex
\usepackage{amsmath,amsfonts,bm}

\def\eqref#1{equation~\ref{#1}}

\def\1{\bm{1}}

\DeclareMathAlphabet{\mathsfit}{\encodingdefault}{\sfdefault}{m}{sl}
\SetMathAlphabet{\mathsfit}{bold}{\encodingdefault}{\sfdefault}{bx}{n}



%% file: intro.tex
\section{Introduction}
Graph analytics is of central importance in many applications including the analysis of social, biological, and financial networks. Many conventional methods focus on measures and metrics that are carefully designed for graphs, and their efficient computation \cite{newman2018networks}. Recently, machine learning based approaches have received great amount of attention, e.g., \cite{tang2015line, grover2016node2vec, wang2016structural, atwood2016diffusion, kipf2016semi, velickovic2017graph, bojchevski2018deep, velivckovic2018deep}, which aim to automatically explore and learn network's structural information. One prominent paradigm is network embedding.  The idea is to learn a latent representation (a.k.a node embedding) for each node in an unsupervised manner, which captures inherent properties and structural information of the underlying graph. Then various downstream tasks can be solved in the latent embedding space with conventional learning and mining tools. 

A successful line of research in network embedding has been inspired by the skip-gram model from the NLP community \cite{mikolov2013distributed}. Deepwalk \cite{perozzi2014deepwalk} and node2vec \cite{grover2016node2vec} first generate a set of paths using various random walk models, then treat these paths as if they were sentences and use them to train a skip-gram model. Instead of random walks, LINE \cite{tang2015line} directly uses the first and second order proximity between nodes as positive examples. The intuition behind these methods is to make sure nodes "close" in the graph are also close in the  embedding space. The key to such methods is how to measure the "closeness" between nodes. 

Recently, there has been great successes in generalizing convolutional neural networks (CNNs) to graph data, e.g., \citep{bruna2013spectral,duvenaud2015convolutional}. Graph convolutional networks (GCNs) proposed by Kipf and Welling \cite{kipf2016semi} and their variants achieve state-of-the-art results in many tasks, which are considered as  better encoder models for graph data. However, most successful methods in this regime use (semi-)supervised training and require lots of labeled data. The performance of GCNs degrades quickly
as the train set shrinks \cite{li2018deeper}. Thus, it is desirable to design appropriate unsupervised objectives that are compatible with convolution based models. 

Hamilton et al. \cite{hamilton2017inductive} suggest to use similar contrastive objectives as in skip-gram to train GCN and its extensions, which is shown to be empirically effective. A recent work \cite{velivckovic2018deep} proposes a more complicated objective function based on mutual information estimation and maximization \cite{belghazi18mutual, hjelm2018learning}, which achieves highly competitive results even compared to state-of-the-art semi-supervised models. This objective is also contrastive, but instead of measuring distance between node pairs, it measures the similarities between single nodes with respect to the representation of the entire graph. Although achieving better performance, the model is more complicated and the computational costs are higher than simpler contrastive objectives, as mutual information estimation and graph-level representations are used in the training process. 

\subsection{Our Method}
In this paper, we provide a new approach called \emph{Sparsest Cut network Embedding}  (\emph{\ourSimpleModel}) for unsupervised network embedding based on well-known graph partition problems. The resulting model is simple, effective and easy to train, often outperforming both semi-supervised and unsupervised baselines in terms of accuracy and computation time. 

\noindent\textbf{Contributions.} First, we propose a novel contrastive-type optimization formulation for network embedding, which draws inspiration from the classic sparsest cut problem. We then propose to use Laplacian smoothing operators (or filters) to simplify the objective function, which aims to eliminate the need for positive examples. As in prior works, a small random sample of node pairs is used as negative examples, but in our framework, we show that this is theoretically justified by spectral graph sparsification theory \cite{spielman2011graph}.
We use graph convolutional networks to implement Laplacian smoothing operators, since it has been proved by Wu et al. \cite{wu2019simplifying} that graph convolutions behavior much like low-pass filters on graph signals.  In a nutshell, the resulting learning model consists of a GCN structure as the encoder and a loss function that only involves negative samples. Finally, extensive empirical studies are conducted, which clearly demonstrate the advantages of our new model in both accuracy and scalability. 

Note that we only use negative samples to train the graph convolutional network, which may seem counter-intuitive.  The main reason why it works is that the iterative neighborhood aggregation schemes in GCN implicitly forces nearby nodes in the graph to be close in the embedding space. Thus, explicit use of positive examples in the objective function could be redundant.  A critical detail in implementation of previous methods is how to draw positive samples. Change of sampling schemes may lead to dramatic drop in performance. Moreover, effective schemes such as random walk based method introduce extra computation burden. On the contrary, there is no need for positive samples in our approach, which not only makes the implementation and tuning much easier, but also reduces the training time significantly.  

\subsection{Preliminaries}
\paragraph{Graph notations}
We will consider an undirected graph denoted as $G=(V, E, F)$, where $V$ is the vertex set, $E$ is the edge set, and $F\in \real^{n\times f}$ is the feature matrix (i.e., the $i$-th row of $F$ is the feature vector of node $v_i$). Let $n=|V|$ and $m=|E|$ be the number of vertices and edges respectively. We use $A\in\{0,1\}^{n\times n}$ to denote the adjacency matrix of $G$, i.e., the $(i,j)$-th entry in $A$ is $1$ if and only if their is an edge between $v_i$ and $v_j$. The degree of a node $v_i$, denoted as $d_i$, is the number of edges incident on $v_i$. The degree matrix $D$ is a diagonal matrix and the its $i$-th diagonal entry is $d_i$.
\paragraph{Laplacian matrix} The Laplacian matrix of a graph $G$ is defined as $L_G=D-A$. A useful property of $L$ is that its quadratic form measures the "smoothness" of a vector with respect to the graph structure. More formally, for any vector $x\in \real^n$, it is easy to verify that
\begin{equation}\label{eqn:LPq}
x^TL_Gx = \frac{1}{2}\sum_{i,j} A_{ij} (x_i-x_j)^2 = \sum_{(v_i,v_j) \in E} (x_i-x_j)^2.
\end{equation}
This can be extended to multi-dimensional cases. For any matrix $X\in \real^{n\times d}$, let $X_i$ be the $i$-th row of $X$, then we have

\begin{equation}\label{eqn:trace}
\mathsf{Tr} (X^TL_GX) = \frac{1}{2}\sum_{i,j} A_{ij} \|X_i-X_j\|^2 = \sum_{(v_i,v_j) \in E}  \|X_i-X_j\|^2.
\end{equation}
Here $\mathsf{Tr(\cdot)}$ is the trace function, i.e., the sum of the diagonal entries, and $\|\cdot\|$ is the Euclidean norm.

\paragraph{Graph cuts}
Let $S\subset V$ be a subset of vertices and $\overline{S} = V \setminus S$ be its complement. We use $E(S,\overline{S})$ to denote the set of crossing edges between $S$ and $\overline{S}$, which are also called \emph{cut edges} induced by $S$. One task in many applications is to partition the graph into two or more disjoint parts such that the number of crossing edges is minimized. When studying such graph partition problems, it is useful to investigate the laplacian matrix of the graph, because the laplacian quadratic form applied on the indicator vector of $S$ is exactly the cut size. Let $x_{S} \in \{0,1\}^n$ be the indicator vector of set $S$, i.e., the $i$-th entry is:
\[
x_{S,i} = \left\{
\begin{array}{lr}
1, & \text{if } v_i\in S \\
0, & \text{otherwise }
\end{array} \right. .
\]
Then it follows from (\ref{eqn:LPq}) that
\begin{equation}\label{eqn:cutsize}
x^T_S L_Gx_S =  \sum_{(v_i,v_j) \in E} (x_{S,i}-x_{S,j})^2 = |E(S,\overline{S})|.
\end{equation}

%% file: Analysis.tex
\section{Theoretical Motivation and Analysis}
In this section, we derive our model \ourSimpleModel~ and discuss its theoretical connections to graph partition problems. 
\subsection{Sparsest Cut} Our model is motivated by the \emph{Sparsest Cut} problem (see e.g., \cite{arora2010logn}). The goal in this problem is to partition the graph into two disjoint parts such that the cut size is small; but we also wish the two disjoint parts to be balanced in terms of their sizes. For the standard minimum cut problem, when there are low degree nodes, min-cuts are likely to be formed by a single node, which are not very useful cuts in typical applications. The balance requirement in sparsest cut problems avoids such trivial cuts and could produce more interesting solutions. Next, we define the problem more formally.

For any subset of nodes $S$, its \emph{edge expansion} is defined as
\begin{equation*}
\phi(S) = \frac{|E(S,\overline{S})|}{\min (|S|,|\overline{S}|)}. 
\end{equation*}
The sparsest cut problem asks to find a set $S^*$ with smallest possible edge expansion. We define  $\phi(G) = \min_{S\subset V}  \frac{|E(S,\overline{S})|}{\min (|S|,|\overline{S}|)}$. In this paper, we consider a slight variant of the above definition, which is to find a set $S^*$ with smallest $\phi'(S^*)$, where
\begin{equation*}
\phi'(S) =   \frac{|E(S,\overline{S})|}{|S||\overline{S}|}.
\end{equation*}
Similarly, we define $\phi'(G) = \min_{S\subset V}  \frac{|E(S,\overline{S})|}{|S||\overline{S}|}$.

The above two sparsest cut formulations are equivalent up to an approximation factor of $2$ because
$$\frac{|S||\overline{S}|}{n} \le \min (|S|,|\overline{S}|) \le \frac{2|S||\overline{S}|}{n}.$$
Unfortunately, both of the above two formulations are NP-hard \cite{matula1990sparsest}. Hence most researches focus on designing efficient approximation algorithms. Currently, algorithms based on \emph{SDP relaxations} achieve the best theoretical guarantee in terms of approximation ratio \cite{arora2010logn}.

\subsection{A Parameterized Relaxation}
Another way to represent edge expansion is 
$$\phi'(S) = \frac{2x^\top _S L_Gx_S }{\sum_{i=1}^n \sum_{j=1}^n (x_{S,i}-x_{S,j})^2 } .$$
Indeed, the numerator $x^\top _S L_Gx_S = |E(S,\overline{S})|$ by (\ref{eqn:cutsize}).  From (\ref{eqn:LPq}), the denominator is exactly $2x^\top _S L_Kx_S $, where $K$ is the \emph{complete graph} defined on the same vertex set as $G$ and $L_K$ is the corresponding Lapalacian matrix.  This is twice the cut size of $(S, \overline{S})$ on the complete graph ( by (\ref{eqn:cutsize})), which is exactly $2|S||\overline{S}|$. Therefore, we have
$$\phi'(G) =  \min_{S\in V} \frac{x^\top _S L_Gx_S }{x_S^\top  L_Kx_S } = \min_{x\in \{0,1\}^n} \frac{x^\top  L_Gx }{x^\top  L_Kx }. $$

This algebraic formulation is still intractable, mainly due to the integral constraints ${x \in \{0,1\}^n}$. It is thus natural to relax these and only require each $x_i \in [0,1]$. More powerful relaxations usually lift each $x_i$ to high-dimensions and consider the optimization problem $$\min_{X \in \real^{n\times d}} \frac{\operatorname{Tr}(X^\top  L_GX) }{ \operatorname{Tr}(X^\top  L_KX) },$$ or equivalently (by (\ref{eqn:trace}))
$$ \min_{X \in \real^{n\times d}} \frac{\sum_{(v_i,v_j) \in E}  \|X_i-X_j\|^2}{\sum_{i=1}^n \sum_{j=1}^n \|X_i-X_j\|^2}.$$
Here $X_i$ can be viewed as a $d$-dimensional embedding of node $v_i$.

The drawback of the above relaxation is that it doesn't utilize node features, which usually contains important information. In this paper, we propose the following relaxation:
\begin{equation}\label{eqn:opt}
\min_{\theta} \frac{\sum_{(v_i,v_j) \in E}  \|g_{\theta}(F_i)-g_{\theta}(F_j)\|^2}{\sum_{i=1}^n \sum_{j=1}^n \|g_{\theta}(F_i)-g_{\theta}(F_j)\|^2}.
\end{equation}
Here $g_{\theta}(\cdot): \real^{f} \rightarrow \real^d$ is a parameterized function mapping feature vectors to $d$-dimensional embeddings, and $F_i$ is the feature of node $v_i$. The goal is to find optimal parameters $\theta$ such that the objective function is minimized.

\subsection{Approximation with Graph Convolutional Networks}
The optimization problem (\ref{eqn:opt}) is highly nonconvex, especially when $g_{\theta}()$ is modeled by neural networks, and thus could be very difficult to optimize. In this section, we provide effective heuristics based on Graph Convolution operations \cite{kipf2016semi} to facilitate the optimization.

The problem (\ref{eqn:opt}) can be viewed as a contrastive game between two players: the denominator wants to maximize pair-wise distances between all pairs, while the numerator tries to make neighboring pairs close. To simplify the problem, we model the behavior of the numerator player by a Laplacian smoothing filter and then remove the numerator from the objective function. Let $\Pi_G$ be a smoothing matrix. For a signal $x$, the value of $x^\top L_Gx$ become smaller after applying a smoothing operater on it. Thus, the objective of the numerator player is implicitly encoded in $\Pi_G$ and will be removed from (\ref{eqn:opt}), which greatly eases the optimization process. 

This trick can also be motivated by stochastic optimization. To minimize objective (\ref{eqn:opt}), the algorithm randomly sample some positive examples and negative examples from $E$ and $V\times V$ respectively in each round, then performs mini-batch update via gradient descent. Consider a step of gradient updates from positive samples. This essentially reduces the value of the Laplacian quadratic form, hence performing smoothing.  Instead of perform gradient updates, our method directly applies a low-pass filter to smooth the signal. 

Let $g_{\theta}(F) \in \real^{n\times d}$ denote the matrix whose $i$-th row is $g_{\theta}(F_i)$, and $\Pi_G g_{\theta}(F) \in  \real^{n\times d}$ be the matrix after smoothing, which contains the embeddings of all nodes. Let $z_i$ denote the embedding of $v_i$, which is the $i$-th row in $\Pi_G g_{\theta}(F)$, and $Z= \Pi_G g_{\theta}(F)$ be the output embedding matrix.
Our loss will be of the form 
\begin{equation}\label{eqn:lossall}
L=\frac{2}{\sum_{i=1}^n \sum_{j=1}^n \|z_i-z_j\|^2} =  \frac{1}{\operatorname{Tr} (Z^\top L_K Z)}.
\end{equation}

It is observed in \cite{li2018deeper} that the graph convolution operation from \cite{kipf2016semi} is a special form of Laplacian smoothing \cite{taubin1995signal}. Moreover, it is proved in \cite{wu2019simplifying} that the effect of iteratively applying this graph convolution is similar to a low-pass-type filter, which projects signals onto the space spanned by low eigenvectors approximately. This is exactly what we need for $\Pi_G$. Therefore, in our model, we implement $\Pi_G$ as a multilayer graph convolution network. Since we will use a multilayer linear network to model $g_{\theta}(\cdot)$, our network structure, i.e. $\Pi_Gg_{\theta}(F)$, is similar to SGC from \cite{wu2019simplifying}. See section \ref{sec:model} for the details. 

\subsection{Negative Sampling and Spectral Sparsification}
One disadvantage of the above loss function is that it contains $n^2$ terms, and thus too time consuming even just to make one pass over them. Thus, we will only randomly sample a small set of pairs $\mathcal{N} \subset V\times V$ in the beginning, which are called \emph{negative samples}. We use $H =(V,\mathcal{N})$ to denote the graph with edge set $\mathcal{N}$ and $L_H$ be its Laplacian. Then the loss becomes
\begin{equation}\label{eqn:lossSparse}
L'=\frac{1}{\sum_{(i, j)\in \mathcal{N}} \|z_i-z_j\|^2} = \frac{1}{\operatorname{Tr} (Z^\top L_H Z)}.
\end{equation}

In fact, well-know graph sparsification results show that minimizing $L'$ is almost equivalent to minimizing $L$.
More specifically, if we sample each possible pair independently with probability $p$, then the spectral sparsification theorem from \cite{spielman2011graph} claims that $x^\top L_Kx \approx x^\top L_Hx/p $ holds for all $x\in \real^n$ simultaneously with high probability provided that the number of sampled edges is $\Theta(n\log n)$ in expectation (or $\Theta(\log n)$ per node). By this result, the two loss functions (\ref{eqn:lossall}) and (\ref{eqn:lossSparse}) are approximately equivalent. Assume $Z$ is the optimal embedding for $L'$, then its loss with respect to $L$ is
$$L(Z) \approx \frac{L'(Z)}{p}\le \frac{L'(Z^*)}{p} \approx L(Z^*).$$
Here $Z^*$ is the optimal embedding with respect to $L$ and the inequality follows from the optimality of $Z$ for $L'$.
We refer to \cite{spielman2011graph} for the quantitative bounds on graph sparsification. 

\subsection{Remarks}
The sparsest cut problem considered above is usually called uniform sparsest cut, which can be formulated as $\min_{x\in \{0,1\}^n} \frac{x^\top  L_Gx }{x^\top  L_Kx }$. A natural generalization is to use different a graph, $G'$, rather than complete graph in the denominator and consider the problem $\min_{x\in \{0,1\}^n} \frac{x^\top  L_Gx }{x^\top  L_{G'}x }$. This more general problem is called non-uniform sparsest cut and many graph partition problems are special cases of this formulation. For instance, if there is only one edge $(s, t)$ in $G'$, then this is equivalent to minimum $s$-$t$ cut. This objective is also contrastive and each edge in $G'$ can be viewed as a negative example. Although we show in this work that the simplest choice of complete graph has already achieves impressive results, in general, more prior information can be encoded in $G'$ to further improve the prediction accuracy. 

The sparsest cut problem only considers bi-partitions. However, it can be extended to multi-partitions and hierarchical partitions by applying a top-down recursive partitioning scheme \cite{charikar2017approximate}. It would be interesting to encode such recursive paradigms into network structure.

Our method is inspired from sparsest cut problem, but is not intended to solve it. It is unclear whether graph neural networks could be helpful for solving such graph partition problems.

%% file: model.tex
\section{Implementation details of our Model}\label{sec:model}
In this section, we provide more implementation details of \ourSimpleModel, and an extension of the basic model will also be considered.  
\subsection{Graph Convolutions}
The graph convolutional filters used in our implementation is originated from \cite{kipf2016semi}. Let $A$ be the adjacency matrix of the underlying graph, $D$ be its degree matrix, and $I$ be the identity matrix.  $\tilde{A}$ is defined as the adjacency matrix of the graph after adding self-loops for each node, i.e., $\tilde{A} = A+I$, and similarly $\tilde{D} = D+I$ is the adjusted degree matrix. The graph convolution proposed in \cite{kipf2016semi} is $\tilde{D}^{-1/2}\tilde{A}\tilde{D}^{-1/2}$. 
In this work, we will use a asymmetric variant suggested in \cite{hamilton2017inductive}, which is $ \tilde{D}^{-1}\tilde{A}.$

\begin{algorithm}[h]
\caption{Message Passing of~\ourSimpleModel}\label{alg:SCE}
\begin{algorithmic}[1]
\Require 
Input feature matrix $F$; Graph adjacent matrix with self-loop $\tilde{A}$; Linear network weights $\{W^{(i)}\}_{i=0}^{l}$
\Ensure 
Output embedding matrix $Z$
\State Initialize $F^{(0)}=F$;
\State Compute the degree matrix $\tilde{D}$ by $\tilde{D}_{ii} = \sum_{j}\tilde{A}_{ij}$;
\State $//$ \emph{Feature Smoothing};
\For{$i=1,2,\cdots,k$}
\State $F^{(i)}=\tilde{D}^{-1}\tilde{A}F^{(i-1)}$;
\EndFor
\State $//$ \emph{Linear Network};
\State Initialize $X^{(0)}=F^{(k)}$;
\For{$j=1,2,\cdots,l$}
\State $X^{(j)}=X^{(j-1)}W^{(j)}$;
\EndFor\\
\Return $Z=X^{(l)}$;
\end{algorithmic}
\end{algorithm}

\subsection{Architecture}
The architecture conceptually consists of two components. The first component is a graph convolutional filter used for smoothing, and the second component is a multilayer linear network modeling the parameterized mapping $g_{\theta}(\cdot)$ described in the previous section. The input is a set of node features represented as a matrix $F\in\real^{n \times f}$, in which $n$ is the number of nodes and $f$ is the dimension of the input feature. 

\paragraph{Feature Mapping} We use multilayer linear network to model the feature mapping $g_{\theta}(\cdot)$, i.e.,
$$g_{\theta}(F) = F W^{(1)}\cdots W^{(l)},$$
and $\theta$ consists of all the parameter matrices $W^{(1)},\cdots, W^{(l)}$. In terms of expressive power, multilayer linear networks are the same as single layer ones. However, if the network is trained with gradient-based method, it has been proved that the optimization trajectory of deep linear networks could differ significantly. In particular, there is a form of implicit regularization induced from training deep linear networks with gradient-based optimization \cite{arora2019implicit}. In our experiments, we also observe multilayer structures often perform better. 

\paragraph{Smoothing Matrix}
The basic implementation of the smoothing matrix $\Pi_{G}$ in our model contains $k$ simple iterations. The $i$-th iteration is formulated as
\begin{align}\label{eq: propagation rule}
F^{(i)}=&\tilde{D}^{-1}\tilde{A}F^{(i-1)}, i=1,2,..., k.\notag
\end{align}
More compactly, the smoothing operater can be written as
\begin{equation*}
\Pi_{G}F=(\tilde{D}^{-1}\tilde{A})^{k}F.
\end{equation*}
So the encoder structure in \ourSimpleModel~ is 
$$(\tilde{D}^{-1}\tilde{A})^{k}FW^{(1)}\cdots W^{(l)}.$$
Detailed illustration can be seen in Algorithm~\ref{alg:SCE}. For large data sets, we also propose a mini-batch version to save computation and memory costs, which can be found in Algorithm~\ref{alg:mini-batch SCE}.

\begin{figure}[!htp]
\centering
 \includegraphics[width=7.5cm]{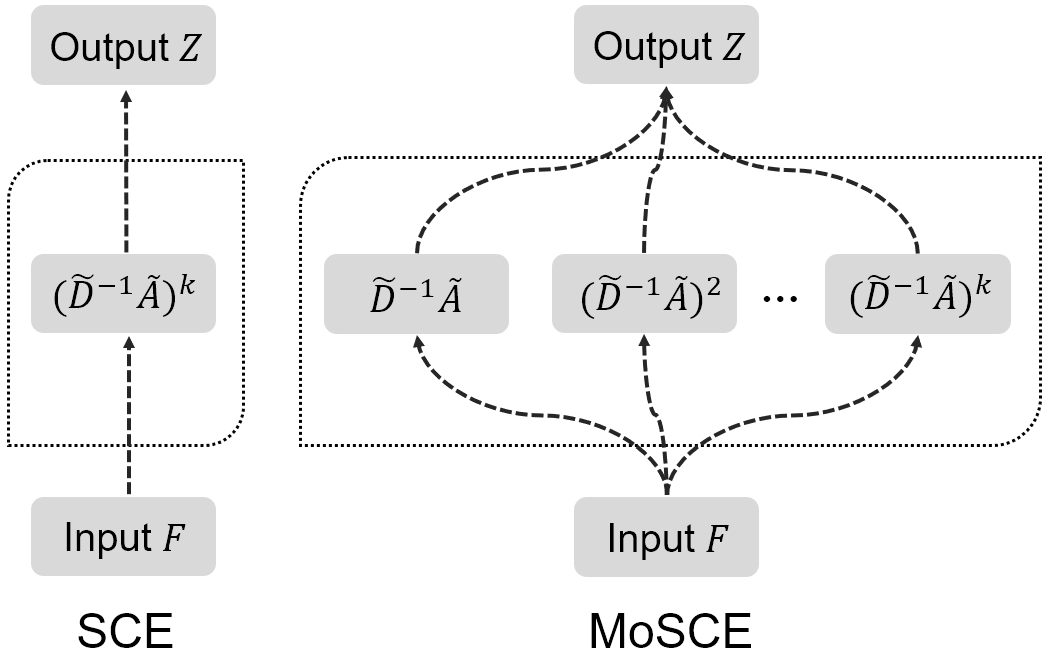}
\caption{The structure of~\ourSimpleModel~and \ourMultiModel}\label{fig:network}
\end{figure}

\begin{algorithm}[h]
\caption{Message Passing of Mini-batch~\ourSimpleModel}\label{alg:mini-batch SCE}
\begin{algorithmic}[1]
\Require 
Input feature matrix $F$; Graph adjacent matrix with self-loop $\tilde{A}$; Linear network weights $\{W^{(i)}\}_{i=0}^{l}$; Mini-batch size $b$;
\Ensure 
Output embedding matrix $Z$
\State Initialize $F_0=F$;
\State Compute the degree matrix $\tilde{D}$ by $\tilde{D}_{ii} = \sum_{j}\tilde{A}_{ij}$;
\State $//$ \emph{Feature Smoothing};
\For{$i=1,2,\cdots,k$}
\State $F^{(i)}=\tilde{D}^{-1}\tilde{A}F^{(i-1)}$;
\EndFor
\State Sample $b$ rows from $F^{(k)}$ and concatenate them into a matrix $\text{Sample}(F^{(k)})$
\State $//$ \emph{Linear Network};
\State Initialize $X_0=\text{Sample}(F^{(k)})$;
\For{$j=1,2,\cdots,l$}
\State $X^{(j)}=X^{(j-1)}W^{(j)}$;
\EndFor\\
\Return $Z=X^{(l)}$;
\end{algorithmic}
\end{algorithm}
\paragraph{\ourMultiModel}
 To aggregate information, we also use multi-scale graph filters using similar ideas as in \cite{abu2019mixhop,abu2018n}.  The model with multi-scale filters is called \emph{Multi-order Sparsest Cut network Embedding}  (\emph{\ourMultiModel}). A graphical illustration of \ourMultiModel~ is presented in Figure~\ref{fig:network}. We compute representations $\{Z^{(i)}\}_{i=1}^{k}$ with different smoothing levels, i.e. $Z^{(i)}=(\tilde{D}^{-1}\tilde{A})^{i}FW^{(1)}\cdots W^{(l_i)}$ and aggregate these smoothed features with concatenation, mean-pooling or max-pooling:
\begin{equation*}
Z=aggregate(Z^{(1)}, Z^{(2)},\cdots, Z^{(k)}).
\end{equation*}

Since $Z^{(1)}, Z^{(2)},\cdots, Z^{(k)}$ are computed in \ourSimpleModel, the computational cost of \ourMultiModel~ is roughly the same.

\subsection{Loss Function}
We use $\theta$ to denote the set of all trainable parameters. With the \emph{negative samples} denoted as $\mathcal{N} \subset V\times V$, the unsupervised loss used for training is
\begin{displaymath}
\mathcal{L}_{unsup}(\theta;A,X)=\frac{1}{\sum_{(v_i, v_j) \in \mathcal{N}}\parallel z_i-z_j\parallel^{2}}.
\end{displaymath}
Here $z_i$ and $z_j$ are the output embeddings of node $v_i$ and $v_j$. Each negative pair in $\mathcal{N}$ is randomly sampled from $V\times V$.
To suppress overfitting, a $L_2$ regularizer is introduced in our loss function as $\mathcal{L}_2=\|\theta\|_2^2$. Thus, the whole loss function of our models is 
$$\mathcal{L}=\alpha \mathcal{L}_{unsup}+\beta\mathcal{L}_{2},$$
where $\alpha$ and $\beta$ are hyper-parameters to balance the effect of regularization. 

%% file: related.tex
\section{Related Work}
 Many methods have been proposed for learning node representations in the past few years. One major class of methods use random walk to generate a set of paths, then treat these paths as if they were sentences and use them to train a skip-gram model. These class includes Deepwalk \cite{perozzi2014deepwalk}, LINE \cite{tang2015line}, node2vec \cite{grover2016node2vec}, VERSE \cite{tsitsulin2018verse} and many others.  Other class is based on matrix factorization \cite{ou2016asymmetric, zhang2018arbitrary}. These methods first (approximately) compute a proximity matrix $M\in \real^{n\times n}$, which represents the pair-wise node similarities, and then use singular value decomposition (SVD) or eigen-decomposition to obtain the desired embeddings. For instance, personalized page rank was used as the proximity \cite{yin2019scalable, zhou2017scalable}. See \cite{tsitsulin2018verse} for more choices of proximity. Recently, it has been shown that many random walk based approaches can be unified into the matrix factorization framework with closed forms \cite{qiu2018network}.  
 
 Other approaches are also widely studied. SDNE \cite{wang2016structural} uses an Autoencoder structure. A recent work of \cite{bojchevski2018deep} proposes Graph2Gauss. This method embeds each node as a Gaussian distribution according to a novel ranking similarity based on the shortest path distances between nodes. With the introduction of graph neural networks, GNN based embedding methods are also widely studied \cite{kipf2016variational, hamilton2017inductive, velivckovic2018deep}, which use unsupervised contrastive-type objective functions.
 
 A critical building of our framework is Laplacian smoothing filters.  In our implementation, we use recent graph convolutional networks \cite{kipf2016semi, wu2019simplifying} for this purpose, since the effect of iteratively applying this graph convolution is similar to a low-pass-type filter, which projects signals onto the space spanned by low eigenvectors approximately \cite{wu2019simplifying, li2018deeper, hoang2019revisiting}. GCNs generalize convolutions to the graph domain, working with a spectral operator depending on the graph structure, which  learns hidden layer representations that encode both graph structure and node features simultaneously \cite{bruna2013spectral,henaff2015deep,defferrard2016convolutional}. Deeper versions of GCN often lead to worse performance \cite{kipf2016semi,xu2018representation}. It is believed that this might be caused by over-smoothing, i.e., some critical feature information may be "washed out" via the iterative averaging process (see e.g., \cite{xu2018representation}). Many approaches are proposed to cope with oversmoothing, e.g., \cite{li2018deeper,xu2018representation, abu2019mixhop,luan2019break}, and such structures are potentially better filter for our framework. 
 
 Our framework draws inspirations from the sparsest cut problem, which is well-known to be NP-hard \cite{matula1990sparsest}. On the other hand, efficient approximation algorithms are known for this problem, most of which are based on Linear Programming or Semi-definite Programming relaxations \cite{leighton1999multicommodity,arora2009expander}.  Currently, algorithms based on \emph{SDP relaxations} achieve $O(\sqrt{\log n})$ approximation, which is conjectured to be the best possible \cite{arora2009expander}. The $O(\log n)$-approximation algorithm from \cite{arora2009expander} has high computational complexity and the time is later improved to $O(n^2 \mathsf{polylog }  (n))$ in \cite{arora2010logn}.

%% file: exp.tex
\section{Experiments}

We evaluate the performance of \ourSimpleModel~ and \ourMultiModel~ on both transductive and inductive node classification tasks. We compare our methods with state-of-the-art unsupervised models, while we also list the results of strong supervised methods for reference. For unsupervised models, the learned representations are evaluated by training and testing a logistic regression classifier. We detail the experimental setup and results in following parts.


\subsection{Experimental Setup}

\begin{table}[t]
\setlength{\abovecaptionskip}{0.2cm}
\setlength{\belowcaptionskip}{0.cm}
	\caption{Summary of the datasets used in our experiments, as reported in \citep{yang2016revisiting, bojchevski2018deep, zeng2020graphsaint} }.	
	\begin{tabular}{lcccc}\toprule \centering
	\textbf{Dataset}&\textbf{Nodes}&\textbf{Edges}&\textbf{Features}&\textbf{Classes}\\
	\midrule
	Cora&2,708&5,429 &1,433&7\\
	Citeseer&3,327&4,732&3,703&6 \\
	Pubmed&19,717&44,338&500&3\\
	Cora Full&19,793&65,311&8,710&70\\
	Flicker&89,250&899,756&500&7\\
	Reddit&232,965&11,606,919&602&41\\
	\bottomrule
	\end{tabular}
	\label{tab:datasets}
	\centering
\end{table}

\textbf{Datasets.} The results are evaluated on four citation networks \citep{kipf2016semi,bojchevski2018deep} Cora, Citeseer, Pubmed, Cora Full for transductive task and two community networks \cite{zeng2020graphsaint} Flicker, Reddit for inductive learning. The citation networks contain sparse bag-of-words feature vectors for each document and a list of citation links between documents. We follow previous literature to construct undirected graphs on them. Models are trained to classify each document (node) to a corresponding label class.  For inductive learning on community networks, the task on Flickr dataset is to categorize images into difference types based on their descriptions and properties, while the task on Reddit is to predict which community each post belongs to. See Table \ref{tab:datasets} for a concise summary of the six datasets.

\textbf{Metrics.} In the literature, models are often tested on fixed data splits for transductive tasks. However, such experimental setup may favors the model that overfits the most \cite{shchur2018pitfalls}. In order to get thorough
empirical evaluation on transductive tasks, results are averaged over 50 random splits for each dataset and standard deviations are also reported. Moreover, we also test the performances under different label rates. Performances for inductive tasks are tested on relatively larger graphs, so we choose fixed data splits as in previous papers \citep{hamilton2017inductive,zeng2020graphsaint} and report the Micro-F1 scores averaged on 10 runs. To evaluate the scalability of different models, we also report their training time.


\textbf{Baseline models.} Baseline models used are list below. We follow the common practice suggested in the original papers for hyperparameter settings.
\begin{itemize}
  \item Deepwalk \cite{perozzi2014deepwalk}: Deepwalk is the representative random walk based method. This method preserves higher-order proximity between nodes by maximizing the probability of the observed random walk.
  
  \item GCN-unsupervised: To demonstrate the effectiveness of our loss function, we include an unsupervised version of GCN. This method simply use a GCN as the encoder and is trained with the same unsupervised loss function used in LINE \cite{tang2015line},  GraphSAGE \cite{hamilton2017inductive}:
\begin{equation}
	L_{u}=-\log ( \sigma (z_{u}^{T}z_{v})) - Q \cdot E_{v_{n} \sim P_{n}} \log ( \sigma (-z_{u}^{T}z_{v_n})). \label{eq:sageloss}
\end{equation}
 Here node $v$ is the neighbor of node $u$ and $P_{n}$ is a negative sampling distribution of node $u$.
 
  \item GraphSAGE \cite{hamilton2017inductive}: GraphSAGE use stochastic training techniques. By applying a neighbor sampling strategy, it can process very large graphs. There are several variants with different aggregators proposed in the paper. The loss function for unsupervised training is (\ref{eq:sageloss}).
  
   \item G2G \cite{bojchevski2018deep}: This method embeds each node as a Gaussian distribution according to a novel ranking similarity based on the shortest path distances between nodes. A distribution embedding naturally captures the uncertainty about the representations.
 
  \item DGI \cite{velivckovic2018deep}: DGI uses GCNs as its encoder and more complicated unsupervised loss for training. The loss function is supposed to maximizes the mutual information between "patch representations and a corresponding summary vector".

\end{itemize}

Methods listed above are unsupervised models, which are the main competitors. We also compare \ourSimpleModel~ with some representative semi-supervised models, namely GCN \cite{kipf2016semi}, SGC \cite{wu2019simplifying}, GAT \cite{velickovic2017graph}, MixHop \cite{abu2019mixhop}.

\textbf{General model setup.} For all the supervised models, we utilize early stopping strategy on each random split and output their corresponding classification result. For all the unsupervised models, we choose embedding dimension to be $512$ on Cora, Citeseer, Cora Full and 256 on Pubmed. After the embeddings of nodes are learned, a classifier is trained by applying logistic regression in the embedding space.
For inductive learning, \ourSimpleModel~ uses 512-dimensional embedding space. Other settings of hyper-parameters for the baseline models are the same as suggested in pervious papers. 

\textbf{Hyperparameter for our models.} 
In the transductive experiments, the detailed hyperparameter settings for Cora, Citeseer, Pubmed, and Cora Full are listed below. 
For \ourSimpleModel, we use Adam optimizer with learning rates of $[0.001, 0.0001, 0.02, 0.01]$ and a $L_2$ regularization with weights $[$5e-4, 1e-3, 5e-4, 5e-4$]$. The number of training epochs are $[20, 200, 50, 20]$. For~\ourMultiModel, we use Adam optimizer with learning rates of $[0.001, 0.0001, 0.02, 0.01]$ and a $L_2$ regularization with weight [5e-4, 1e-3, 0, 0]. The number of training epochs are $[20, 50, 100, 30]$. We sample 5 negative samples for each nodes on each dataset before training, and the hyperparameter $\alpha$ is set to $[15000, 15000, 50000, 100000]$ respectively. For the multi-scale filter in \ourMultiModel, the number of levels used are $3$ for Cora and Pubmed, $2$ for Cora Full and Citeseer.
As different variants of our model produce almost the same results on inductive tasks, we only list results of the basic version. We train \ourSimpleModel~with an Adam optimizer with a learning rate of $0.001$ and a $L_2$ regularization with weight $0.02$; the number of training epochs are $[4, 20]$ for Reddit and Flickr respectively.

\textbf{Implementation details.} Our models are implemented with Python3 and PyTorch, while for other baseline methods, we use the public release with settings of hyper-parameters the same as suggested in original papers. Experiments are mostly conducted on a NVIDIA 1080 Ti GPU. However, for inductive learning, because the official code of Deepwalk \cite{perozzi2014deepwalk} can only run on cpu and due to the huge memory usage, DGI \cite{velivckovic2018deep} cannot run on GPU, training time of all the models for inductive learning is tested on Intel(R) Xeon(R) CPU E5-2650 v4 (48 cores).

\subsection{Results and Analysis}
\begin{table*}[!htbp]
\setlength{\abovecaptionskip}{0.2cm}
\setlength{\belowcaptionskip}{-0.05cm}
	\caption{Summary of results in terms of mean classification accuracy and standard deviation (in percent) over 50 random splits on different datasets. The size of training set are [5, 20] per class for each dataset respectively. The highest accuracy in each column is highlighted in bold and the top 1 unsupervised are underlined. We group all models into three categories: GNN variants(GCN, GAT, SGC, MixHop), unsupervised embedding methods (DeepWalk, GCN-unsupervised, G2G, DGI) and our models.}\label{tab:result1}
	\centering
\begin{tabular}{clcccccccc}\toprule
\multicolumn{2}{c}{\multirow{2}*{\textbf{Method}}}& \multicolumn{2}{c}{\textbf{Cora}} & \multicolumn{2}{c}{\textbf{Citeseer}}& \multicolumn{2}{c}{\textbf{Pubmed}}& \multicolumn{2}{c}{\textbf{Cora Full}} \\ \cmidrule(r){3-10}
 & &5 &20&5 &20&5 &20&5 &20\\ \midrule 
\multirow{4}{*}{\textbf{Supervised}} & GCN &67.5$\pm$4.8&79.4$\pm$1.6&57.7$\pm$4.7&69.4$\pm$1.4& 65.4$\pm$5.2 &\textbf{77.2}$\pm$2.1& 49.3$\pm$1.8 & \textbf{61.5}$\pm$0.5    \\ 
& SGC   &63.9$\pm$5.4&78.3$\pm$1.9 & 59.5$\pm$3.4&69.8$\pm$1.4     &  65.8$\pm$4.4 &  76.3$\pm$2.3&46.0$\pm$2.2          &57.7$\pm$1.2   \\ 
& GAT   &  71.2$\pm$3.5         &79.6$\pm$1.5&54.9$\pm$5.0     &69.1$\pm$1.5     & 65.5$\pm$4.6  & 75.4$\pm$2.3&43.9$\pm$1.5     &56.9$\pm$0.6\\ 
& MixHop &67.9$\pm$5.7&80.0$\pm$1.4  &54.5$\pm$4.3&67.1$\pm$2.0&64.4$\pm$5.6&75.7$\pm$2.7    &47.5$\pm$1.5&61.0$\pm$0.7\\ \midrule 
\multirow{4}{*}{\textbf{Unsupervised}} & Deepwalk  & 60.3  $\pm$4.0             & 70.5 $\pm$1.9 & 38.3$\pm$2.9             &45.6$\pm$2.0     & 60.3 $\pm$5.6  &70.8 $\pm$2.6     &   38.9$\pm$1.4&51.1$\pm$0.7  \\
& GCN(unsup)  &61.3$\pm$4.3   & 74.3$\pm$1.6             & 42.3$\pm$3.4    &56.8$\pm$1.9  &  60.9$\pm$5.7  & 70.3$\pm$2.5 &        32.7$\pm$1.9& 45.2$\pm$0.9\\
& G2G  &72.7 $\pm$2.0  &76.2 $\pm$1.1& 60.7$\pm$3.5           & 65.7$\pm$1.5     & \underline{\textbf{67.6}}$\pm$3.9&74.1$\pm$2.1&38.9$\pm$1.3& 49.3$\pm$0.5  \\
& DGI  & 72.9 $\pm$4.0 & 78.1$\pm$1.8  &65.7$\pm$3.6&\underline{\textbf{71.1}}$\pm$1.1&65.3$\pm$5.7&73.9$\pm$2.3     & 50.5$\pm$1.4          &58.4$\pm$0.6 \\ \midrule
\multirow{2}{*}{\textbf{Ours}}& \ourSimpleModel  &74.3$\pm$2.7&80.2$\pm$1.1  &65.4$\pm$2.9&70.7$\pm$1.2 & 65.0$\pm$4.9&75.8$\pm$2.2     & 51.3$\pm$1.5& 60.6$\pm$0.6   \\
& \ourMultiModel &\underline{\textbf{74.6}}$\pm$2.9&\underline{\textbf{80.4}}$\pm$1.2   &\underline{\textbf{66.1}}$\pm$2.5&70.8$\pm$1.3&64.8$\pm$4.6   &\underline{75.9}$\pm$2.3    &\underline{\textbf{51.7}}$\pm$1.4 & \underline{61.1}$\pm$0.5    \\
\bottomrule
\end{tabular}
\end{table*}

The numerical results are summarized in Table \ref{tab:result1} (Transductive learning) and Table \ref{tab:result2} (Inductive learning).
 
\subsubsection{Transductive Learning}
In this section we consider transductive learning where the information of the whole dataset is available in the training process. 

\textbf{Comparison Between Unsupervised Embedding Baselines and Our Models.} 
We observe that the performance of \ourSimpleModel~ and its extension \ourMultiModel~ are better than other unsupervised models under most experimental settings. In particular, our model outperforms GCN-unsupervised for all four datasets. We also observe in our experiments that GCN-unsupervised is very difficult to train, and in many cases it works better without training. This shows that a carefully designed unsupervised objective is critical to train GCN-type encoders. In addition, our models typically outperform the best baseline DGI by a margin of 1\%-2\%.

To test the scalability of unsupervised models, we also test the training time for \ourSimpleModel, DGI and G2G, which are listed in Table \ref{tab:time}.
The training time of our models is orders of magnitude faster  than DGI and G2G. For \ourSimpleModel, we use a larger number of training epochs on Citeseer and Pubmed, due to slower convergence on these two datasets. Even so, our training time is still less than 3 seconds, which is roughly 10 times faster than DGI. Among baseline methods,  DGI often has both higher accuracy and lower training time in our experiments. 

To better investigate the tradeoff between training time and testing accuracy, we record the testing accuracy after each training epoch, then plot the wall clock time vs accuracy curve. The results of \ourSimpleModel~ and DGI are presented in Figure \ref{fig:time}. As we clearly see from the figure, for all the four datasets, \ourSimpleModel~ converges much faster than DGI does. 

\textbf{Comparison Between semi-supervised GNN Variants and Our Models.}
The results clearly demonstrate the advantage of our models across four datasets. In particular, our models outperform all GNN variants on Cora, Citeseer and Cora Full. When the number of labels per class is $5$, our models outperform GCN by a margin of 10.1\% on Cora, 8.4\% on Citeseer and 2.4\% on Cora Full. Even though \ourSimpleModel~ is slightly worse than \ourMultiModel, it still surpasses GNN variants in most cases. On the one hand, semi-supervised models are greatly affected by the size of the training set, which is also observed in \cite{li2018deeper}, especially for more sophisticated models such as GAT and MixHop. On the other hand, the accuracy of \ourSimpleModel~ and \ourMultiModel~ are not affected as significantly as others when the size of training set become smaller.


\begin{table}[tbp]
\setlength{\abovecaptionskip}{0.2cm}
\setlength{\belowcaptionskip}{0cm}
	\caption{The average training time of each model on 10 runs. We train \ourSimpleModel~for a fixed number of epochs (20 on Cora, 200 on Citeseer, 100 on Pubmed, 20 on Cora Full). Note that G2G and DGI need to use a large amount of memory. In order to compare fairly on the GPU, we tested the time to learn 128-dimensional vectors on Pubmed and Cora Full.}\label{tab:time}
	\centering
	\begin{tabular}{ccccc}\toprule \centering
	\textbf{Method}&\textbf{Cora}&\textbf{Citeseer}&\textbf{Pubmed}&\textbf{Cora Full}\\
	\midrule 
    G2G&451.5s &89.3s &715.6s&491.1s\\
    DGI&15.7s&16.8s&16.3s&548.2s\\
    \ourSimpleModel&\textbf{0.1}s &\textbf{1.4}s &\textbf{2.3}s&\textbf{0.9}s\\
\bottomrule
	\end{tabular}
\end{table}

\begin{figure*}[]
\setlength{\abovecaptionskip}{0.cm}
\setlength{\belowcaptionskip}{0cm}
\centering
\subfigure[Cora]
 {\includegraphics[width=0.24\textwidth]{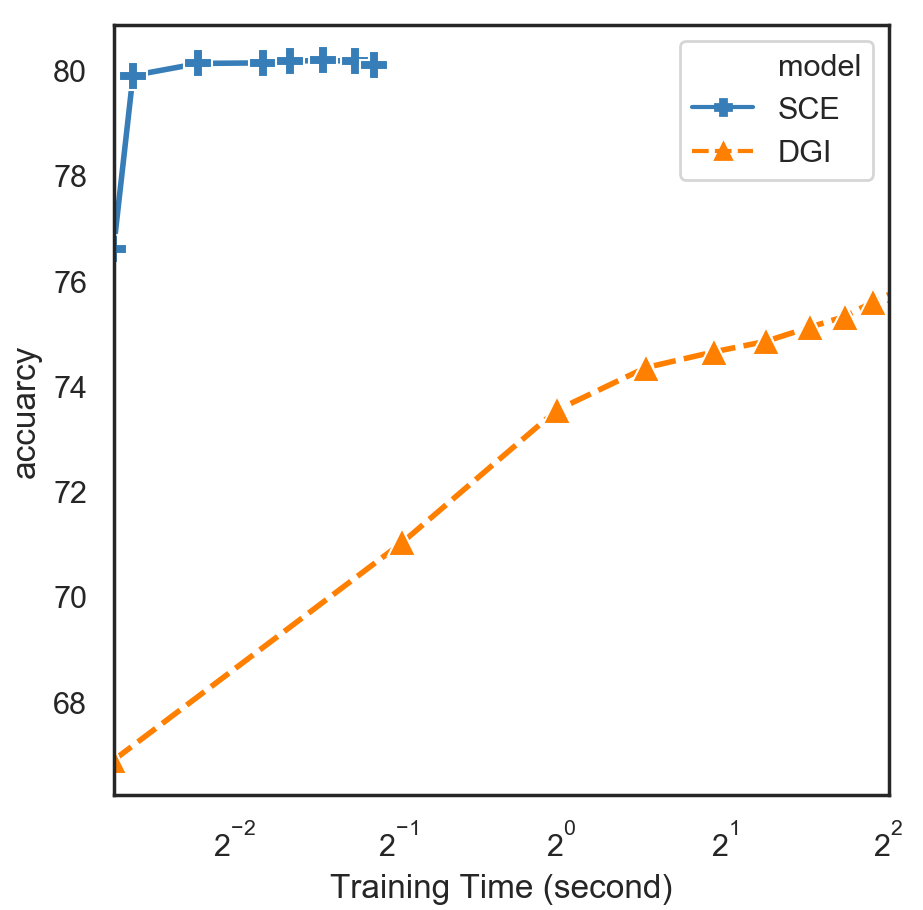}}
\subfigure[Citeseer]
 {\includegraphics[width=0.24\textwidth]{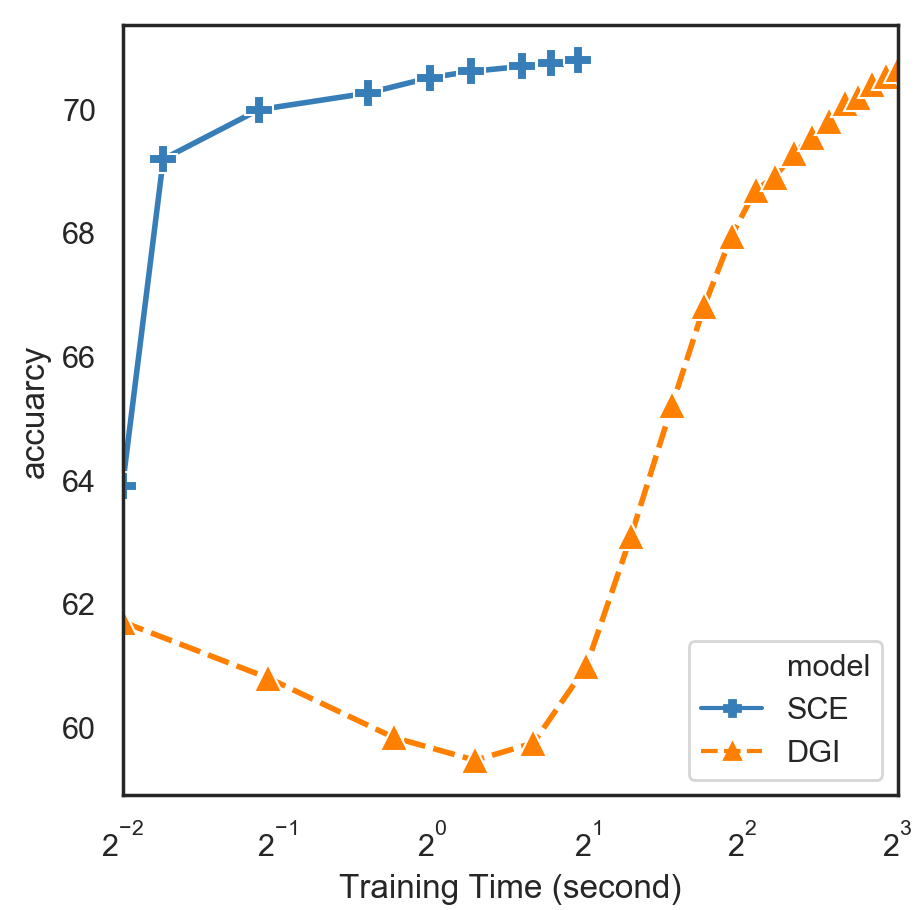}}
\subfigure[Pubmed]
 {\includegraphics[width=0.24\textwidth]{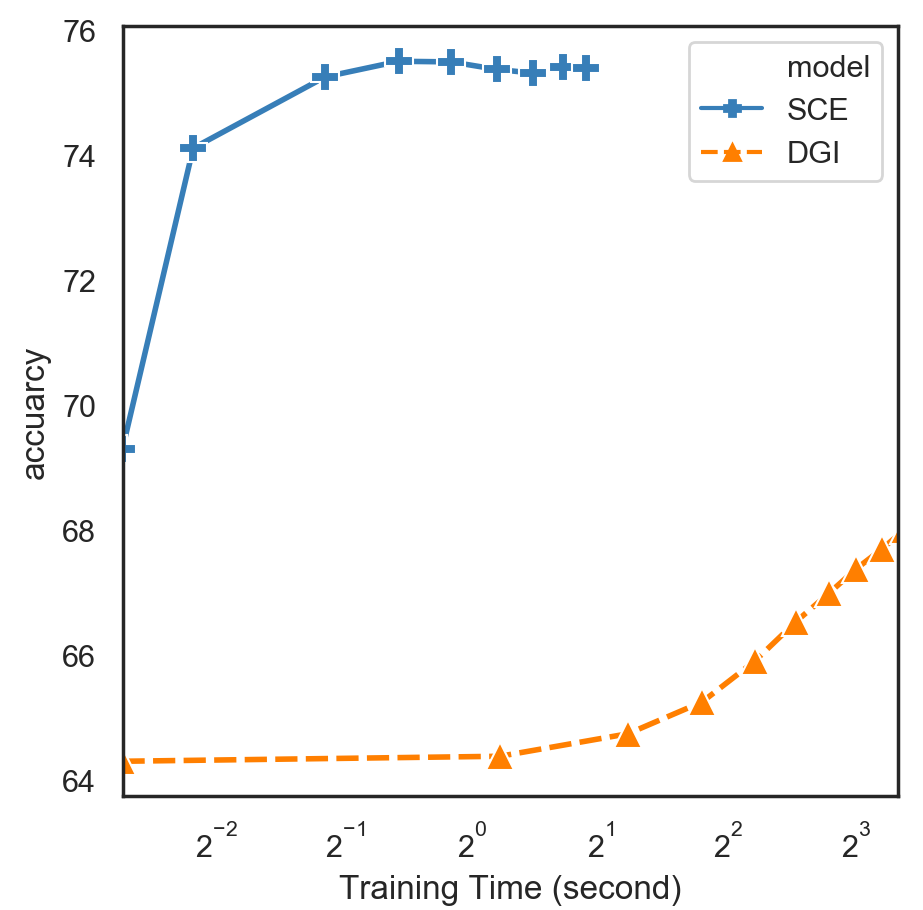}}
 \subfigure[Cora Full]
 {\includegraphics[width=0.24\textwidth]{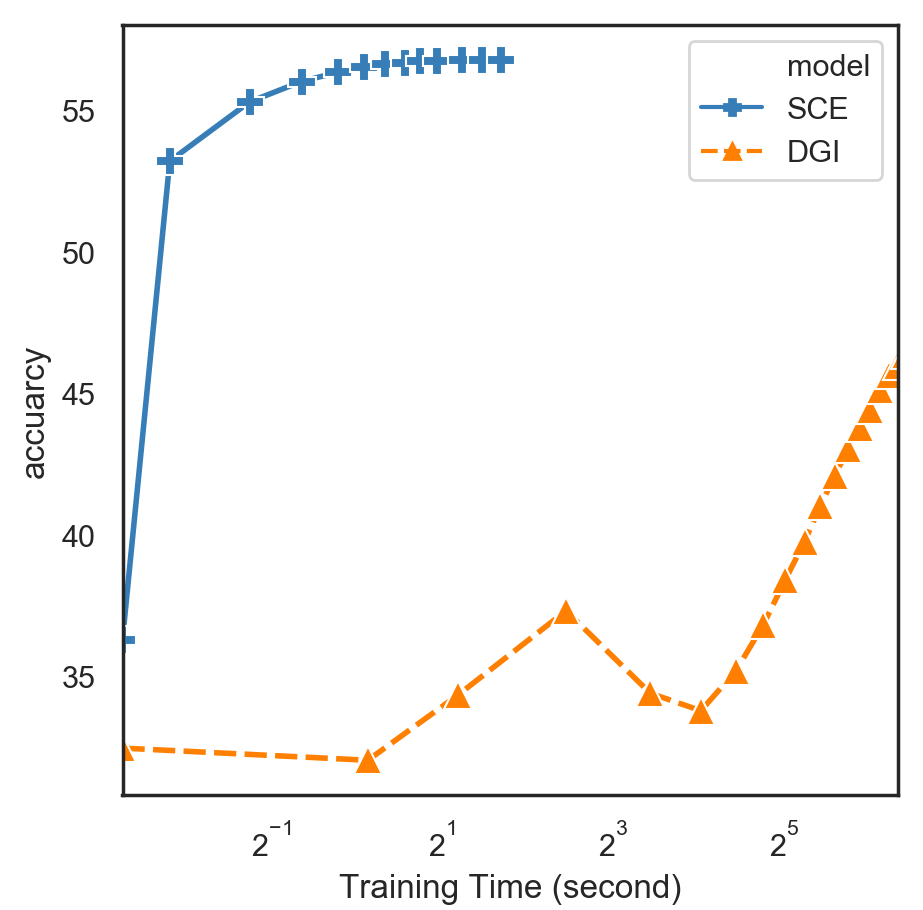}}
  \caption{Test average accuracies on random splits in one training process.}
\label{fig:time}
\end{figure*}

\textbf{Effectiveness of our loss function.}
Here we also study the efficacy of proposed loss function for training graph networks. In our framework, we try to maximize pair-wise distances for negative samples, i.e., $\sum_{(v_i, v_j) \in \mathcal{N}}\parallel z_i-z_j\parallel^{2}$. Our loss function is the inverse of this. Another natural idea is to minimize the negative of these pair-wise distances between negative pairs. So we compare our loss function against the negative sum of Euclidean distances loss:
\begin{displaymath}
\mathcal{L}_{unsup}(\theta;A,X)=-\sum_{(v_i, v_j) \in \mathcal{N}}\parallel z_i-z_j\parallel^{2}
\end{displaymath}

We also test the model accuracy when it is not trained (simply uses randomly initialized parameters).  For these experiments, the encoder structures are the same as in \ourSimpleModel~. 
Table \ref{tab:loss} show the result of different loss functions. Our loss function has a large improvement over both the untrained model and the model trained by the negative Euclidean distance loss. Comparing the negative sum of Euclidean distance loss function with the untrained model, training usually leads to better performance. These results demonstrate the effectiveness of our loss function. 

\begin{table}[tbp]
\setlength{\abovecaptionskip}{0.2cm}
\setlength{\belowcaptionskip}{0cm}
	\caption{The performance of different loss functions.}\label{tab:loss}
	\centering
	\begin{tabular}{lcccc}\toprule \centering
	\textbf{Method}&\textbf{Cora}&\textbf{Citeseer}&\textbf{Pubmed}&\textbf{Cora Full}\\
	\midrule 
	\ourSimpleModel~ (no loss) &76.6$\pm$1.3&63.9$\pm$1.6&   69.4$\pm$2.1  & 51.7$\pm$0.8   \\
	\ourSimpleModel~ (negative)&79.0$\pm$1.5&67.9$\pm$1.6&   65.4$\pm$3.3  & 57.6$\pm$0.6   \\
    \ourSimpleModel~ (ours)& \textbf{80.2}$\pm$1.1    &\textbf{70.7}$\pm$1.2    &   \textbf{75.8}$\pm$2.2 & \textbf{60.6}$\pm$0.6   \\
\bottomrule
	\end{tabular}
\end{table}

\subsubsection{Inductive Learning}
Next we present the results of inductive learning. In this setting, models have no access to test set. The model with higher ability on generalization will perform better on this task.

\textbf{Accuracy.} Experimental results can be found in Table \ref{tab:inductive}, where we only list training time for unsupervised models. 
It's obvious that our model gets a significant performance gain (1\% - 5\% in Micro-F1 scores), producing a score very close to supervised methods \footnote{On large graph dataset like Reddit and Flickr more than 60\% data is in training set, which means supervised method can get much more information than unsupervised methods.}. Meanwhile, it is also much faster (around 1-2 orders of magnitude) than all the previous unsupervised methods.  As stated before, one of the most impressive attributes of our method is its fast training time with limited memory usage.  

DGI uses large amount of memory for processing large graphs, so we were not be able to test the performance of DGI on Reddit under our system condition. The Micro-F1 score of 94.0 is reported in the original paper \cite{velivckovic2018deep}.

 \textbf{Scalability.} Previous unsupervised methods like GraphSAGE and DGI use GCN as their building block, and in order to scale to large graph, they need neighbor sampling strategies under limited GPU memory. Under this space-time trade-off strategy, several redundant forward and backward propagations are conducted on each single node during each epoch, which results in their very long training time.

  However, using a simplified convolution layer as in SGC \cite{wu2019simplifying}, our method does not need message aggregation during training, which tremendously reduces the memory usage and training time. For graphs with more than 100 thousands nodes and 10 millions edges (Reddit), our model can also run smoothly on one NVIDIA 1080 Ti GPU. For even larger graph datasets, stochastic batch training methods are inevitable. For these setting, the simplified convolution layer is also friendly to mini-batch training and does not need complicated subsampling strategy as needed for other models.

\begin{table}[!tbp]\small
\setlength{\abovecaptionskip}{0.2cm}
\setlength{\belowcaptionskip}{0cm}
    \caption{Summary of results in terms of F1 score and runtime on Reddit and Flicker. Because GraphSAGE-LSTM and DGI use too much memory and runtime during training, we cannot reproduce GraphSAGE-LSTM and DGI on Reddit dataset.}\label{tab:result2}
    \begin{tabular}{lllll}
     \toprule
 \multirow{2}{*}{\textbf{Method}} & \multicolumn{2}{c}{\textbf{Reddit}} & \multicolumn{2}{c}{\textbf{Flickr}} \\
 & Micro-F1 & Time & Micro-F1 & Time \\
      \hline
       Raw features & 58.5 & - & 46.2 & - \\
  DeepWalk + features & 69.1 & 1277.9s & 46.1 & 447.5s \\
      \hline
       GraphSAGE-GCN & 90.8 & 477.2s & 43.9 & 29.6s \\
       GraphSAGE-mean & 89.7 & 487.6s & 47.3 & 30.5s \\
       GraphSAGE-LSTM & 90.7 & - & 46.5 &2784.7s \\
       GraphSAGE-pool & 89.2 & 22328.2s & 46.4 &469.0s \\
      DGI & 94.0 & - & 44.7 & 2444.6s  \\
      \hline
       \ourSimpleModel & \underline{\textbf{94.6}} & \underline{\textbf{13.7s}} & \underline{\textbf{50.6}} & \underline{\textbf{16.0s}} \\
   \bottomrule
    \end{tabular}
    \centering
    \label{tab:inductive}
  \end{table}

%% file: conclusion.tex
\section{Conclusion}
In this paper, we provide a new approach for unsupervised network embedding based on graph partition problems. The resulting model is simple, fast and outperforms DGI \cite{velivckovic2018deep} in terms of accuracy and computation time. Our method is based on a novel contrastive objective inspired from the well-known sparsest cut problem. To solve the underlying optimization problem, we introduce a Laplacian smoothing trick, which uses graph convolutional operators as low-pass filters for smoothing node representations. The resulting model consists of a GCN-type structure as the encoder and a simple loss function. Notably, our model does not use positive samples but only negative samples for training, which not only makes the implementation and tuning much easier, but also reduces the training time significantly.   Extensive experimental studies on real world data sets clearly demonstrate the advantages of our new model on both accuracy and scalability.

\textbf{Future work.} The sparsest cut problem considered in the paper is called the uniform sparsest cut. A natural generalization is to use different a graph, $G'$, rather than complete graph, as the negative sample graph and consider the problem $\min_{x\in \{0,1\}^n} \frac{x^T L_Gx }{x^T L_{G'}x }$. We show in this work that the simplest choice of complete graph has already achieves impressive results. We believe more prior information can be encoded in $G'$ to further improve the embedding performance.  The sparsest cut problem only considers bi-partitions. However, it can be extended to multi-partitions and hierarchical partitions by applying a top-down recursive partitioning scheme \cite{charikar2017approximate}. It would be interesting to encode such recursive paradigms into network structure.

\section{Acknowledgments}
This work is supported by Shanghai Science and Technology Commission Grant No. 17JC1420200, National Natural Science Foundation of China Grant
No. 61802069, Science and Technology Commission of Shanghai Municipality Project Grant No. 19511120700,
 and by Shanghai Sailing Program Grant No. 18YF1401200.